\documentclass{article}
\usepackage{ijcai16}
\usepackage{times}
\usepackage{helvet}
\usepackage{courier}
\usepackage{amsfonts,amssymb}
\usepackage{amsmath}
\usepackage{graphicx}
\usepackage{float}
\usepackage{array}
\usepackage{multirow}
\usepackage{url}

\title{From One Point to A Manifold: \\ Knowledge Graph Embedding For Precise Link Prediction}
\author{Han Xiao$^*$, Minlie Huang$^*$, Xiaoyan Zhu \\
	State Key Lab. of Intelligent Technology and Systems, \\
	National Lab. for Information Science and Technology, \\ Dept. of Computer Science and Technology, Tsinghua University, Beijing 100084, PR China
	\\ bookman@vip.163.com; \{aihuang,zxy-dcs\}@tsinghua.edu.cn \\
	$^*$Corresponding Authors: \url{http://www.ibookman.net}, \url{http://www.aihuang.org}
}

\begin{document}

\maketitle

\begin{abstract}
Knowledge graph embedding aims at offering a numerical knowledge representation paradigm by transforming the entities and relations into continuous vector space. However, existing methods could not characterize the knowledge graph in a fine degree to make a precise link prediction. There are two reasons for this issue: being an ill-posed algebraic system and adopting an overstrict geometric form. As precise link prediction is critical for knowledge graph embedding, we propose a manifold-based embedding principle (\textbf{ManifoldE}) which could be treated as a well-posed algebraic system that expands point-wise modeling in current models to manifold-wise modeling. Extensive experiments show that the proposed models achieve substantial improvements against the state-of-the-art baselines, particularly for the precise prediction task, and yet maintain high efficiency. All of the related poster, slides, datasets and codes have been published in \url{http://www.ibookman.net/conference.html}.
\end{abstract}

\section{Introduction}
%Knowledge is critical to intelligence and the numerical representations of knowledge take advantage of numerical machine learning methods to make knowledge computation more generalized and efficient. Specifically, knowledge graph embedding as a numerical representation method is proposed and this approach projects the entities and relations into a continuous high-dimension space then applies a score function to measure the plausibility of the proposal facts. By this mean, knowledge graph completion and other related tasks are promoted.

\begin{figure*}
	\centering
	\includegraphics[width=0.7\linewidth]{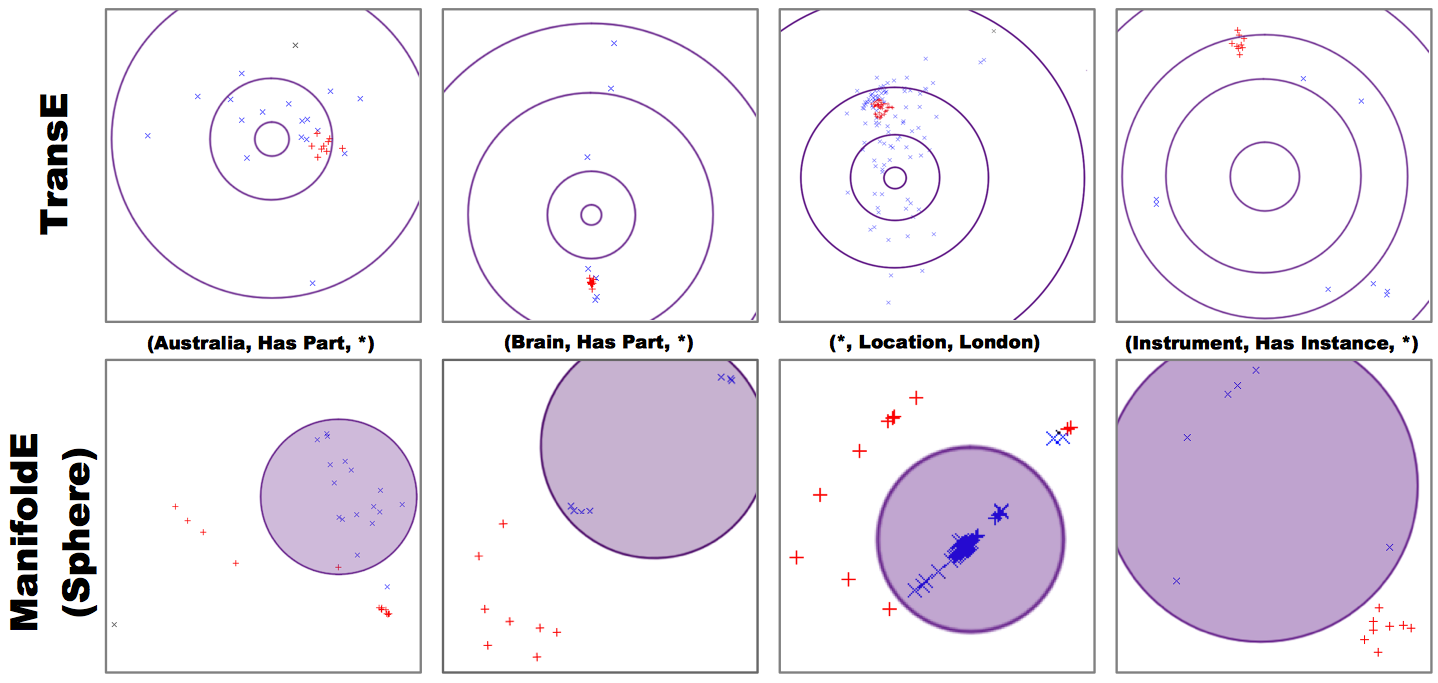}
	\caption{The visualization comparison of TransE and ManifoldE (Sphere). For ManifoldE, the manifold collapses to a solid circle by dimension reduction. The data are selected from Wordnet and Freebase. The blue crosses mean the correctly matched entities while the red crosses indicate the unmatched ones. The upper block corresponds to TransE where more near to the center , more plausible the triple is. It is clear that the true and false triples near the golden position is chaotically distributed. The below block is ManifoldE (Sphere) where the triples inside the solid circle are matched and those outside are unmatched. We can see that there are relatively less errors in ManifoldE than TransE.}
	\label{fig:fig_1}
\end{figure*}

Knowledge is critical to artificial intelligence, and the embedded representation of knowledge offers an efficient basis of computing over symbolic knowledge facts. More specifically, knowledge graph embedding projects entities and relations into a continuous high-dimension vector space by optimizing well-defined objective functions. A variety of methods have been proposed for this task, including TransE \cite{bordes2013translating}, PTransE \cite{lin2015modeling} and KG2E \cite{he2015learning}.

%Diving into the prior details, the triple $(h,r,t)$ of knowledge graph, which indicate the head entity, the relation and the tail entity, respectively, is represented as numerical vectors by embedding. Translation-based embedding methods, such as TransE \cite{bordes2013translating}, PTransE \cite{lin2015modeling} and KG2E \cite{he2015learning}, treat the triple as a relation-specific translation from head to tail entity. Notably, translation-based methods are the key branch of embedding, yielding the state-of-the-art predictive performance.

A fact of knowledge graph is usually represented by a triple $(h,r,t)$,  where $h,r,t$ indicate the head entity, the relation and the tail entity, respectively. The goal of knowledge graph embedding is to obtain the vectorial representations of triples, i.e., $\mathbf{(h,r,t)}$, with some well-defined objective functions. As a key branch of embedding methods, translation-based methods, such as TransE \cite{bordes2013translating}, PTransE \cite{lin2015modeling} and KG2E \cite{he2015learning}, treat the triple as a relation-specific translation from the head entity to the tail entity, or formally as $\mathbf{h+r=t}$.

%Despite the achievement of previous methods, \textit{the \textbf{precise link prediction}, which definitively provides only one answer entity for a query fact}, is also challenging. Currently, for a specific query fact, most existing methods would extract a few candidate entities that may contain a right one, but predicting only one exact entity for a query fact would still make an urgent task. Generally, precise link prediction would improve the feasibility of knowledge completion, the effectiveness of knowledge reasoning and the performance of many knowledge-related tasks. For an instance of knowledge completion, when we want to know about the birth place of Martin R.R., what we desire is only one right answer, while a few answer candidates make not much sense.  Again, for an example of IR, after we extract the query fact from the query text, one unique answer is more preferred than a list, making precise link prediction more valuable to explore.

Despite the success of previous methods, none of previous studies has addressed the issue of \textit{\textbf{precise link prediction}, which finds the exact entity given another entity and the relation}. For a specific query fact, most existing methods would extract a few candidate entities that may contain correct answers, but there is no mechanism to ensure that the correct answers rank ahead the candidate list.

Generally speaking, precise link prediction would improve the feasibility of knowledge completion, the effectiveness of knowledge reasoning, and the performance of many knowledge-related tasks. Taking knowledge completion as example, when we want to know about the birth place of Martin R.R., what we expect is the exact answer ``U.S.'', while a few other candidates do not make any sense.

The issue of precise link prediction is caused by two reasons: \textit{the ill-posed algebraic system} and \textit{the over-strict geometric form}.

First, from the algebraic perspective, each fact can be treated as an equation of $\mathbf{h_r+r=t_r}$\footnote{More generally speaking, $\mathbf{h_r}$ and $\mathbf{t_r}$ are embedding vectors projected w.r.t the relation space, and $\mathbf{r}$ is the relation embedding vector.} if following the translation-based principle, embedding could be treated as a solution to the equation group. In current embedding methods, the number of equations is more than that of free variables, which is called an ill-posed algebraic problem as defined in \cite{Tikhonov1978Solutions}. More specifically, $\mathbf{h_r+r=t_r}$ indicates $d$ equations as $\mathbf{h_i+r_i=t_i}$ where $d$ is the dimension of embedding vector and $i$ denotes each dimension. Therefore, there are $T*d$ equations where $T$ is the number of facts, while the number of variables are $(E+R)*d$, where $E,R$ are the number of entities and relations, respectively. As is the case that triples are much more than the sum of entities and relations, the number of variables are much less than the number of equations, which is typically an ill-posed algebraic system. Mathematically, an ill-posed algebraic system would commonly make the solutions imprecise and unstable. In this paper, we propose to address this issue by replacing the translation-based principle $\mathbf{h_r+r=t_r}$ by a  manifold-based principle $\mathbf{\mathcal{M}(h, r, t)} = D_r^2$ where $\mathcal{M}$ is the manifold function. With the manifold-based principle, our model can make a nearly well-posed algebraic system by taking $d\geq \frac{T}{E+R}$ so that
the number of equations ($T$) is no more than that of the free parameters ($(E+R)*d$).

Second, from the geometric perspective, the position of the golden facts in existing methods is almost one point, which is too strict for all relations and is more insufficient for complex relations such as many-to-many relations. For example, for entity \textit{American Revolution}, there exist many triples such as \textrm{(American~Revolution, Has~Part, Battle~Bunker~Hill)}, \textrm{(American~Revolution, Has~Part, Battle~Cowpens)}. When many tail entities compete for only one point, there would be a major loss of objective function. Some previous work such as TransH \cite{wang2014knowledge} and TransR \cite{lin2015learning} address this problem by projecting entities and relations into some relation-specific subspaces. However, in each subspace, the golden position is also one point and the over-strict geometric form is still existing. As can be seen from Fig.\ref{fig:fig_1}, the translation-based geometric principle involves too much noise. However, ManifoldE alleviates this issue by expanding the position of golden triples from one point to a manifold such as a high-dimensional sphere. By this mean, ManifoldE avoids much noise to distinguish the true facts from the most possible false ones, and improves the precision of knowledge embedding as Fig.\ref{fig:fig_1} shows.

To summarize, our contributions are two-fold:
\textbf{(1)}We have addressed the issue of \textbf{\textit{precise link prediction}} and uncover the two reasons: \textit{the ill-posed algebraic system} and \textit{the over-strict geometric form}. To our best knowledge, this is the first time to address this issue formally.
\textbf{(2)}We propose a manifold-based principle to alleviate this issue and design a new model, ManifoldE, which achieves remarkable improvements over the state-of-the-art baselines in experiments, particularly for precise link prediction. Besides, our methods are also \textit{very efficient}. %Notably, our work is comparable to other related methods in all the tasks, and strong in precise link prediction.

%\begin{figure}
%\centering
%\includegraphics[width=1.0\linewidth]{./fig_2}
%\caption{Visualization comparasion between translation-based principle and manifold-based princile. The solid means the matched entities and the hollow means unmathched ones. Translation-based principle triggers the unsatisfactory precise prediction issue and involves much noise by fitting with a big loss, as (a), while manifold-based principle solves this issue to avoid much noise by more precisely fitting the knowledge graph, as (b).}
%\label{fig:fig_2}
%\end{figure}

%The paper is organized as followed, we survey the related work in Section 2 and then introduce our method in Section 3. The experiments are followed in Section 4. Finally, we conclude our work in Section 5.

\section{Related Work}
%There are two key branches of knowledge graph embedding: translation-based methods and others.

\subsection{Translation-Based Methods}
As a pioneering work of knowledge graph embedding, \textbf{TransE} \cite{bordes2013translating} opens a line of translation-based methods. TransE treats a triple $(h,r,t)$ as a relation-specific translation from a head entity to a tail entity, say $\mathbf{h+r = t}$ and the score function has a form of $||\mathbf{h+r-t}||_2^2$. Following this principle, a number of models have been proposed.
For instance, \textbf{TransH} \cite{wang2014knowledge} adopts a projection transformation, say $\mathbf{h_r=h-w_r^\top h w_r}$, $\mathbf{t_r=t-w_r^\top t w_r}$, while \textbf{TransR} \cite{lin2015learning} applies a rotation transformation, say $\mathbf{h_r=M_rh}$,$ \mathbf{t_r=M_rt}$.
Similar work also includes \textbf{TransD} \cite{JiKnowledge} and \textbf{TransM} \cite{fan2014transition}. Other approaches take into consideration extra information such as relation-type \textbf{\cite{wang2015knowledge}}, paths with different confidence levels (\textbf{PTransE}) \cite{lin2015modeling}, and semantic smoothness of the embedding space \textbf{\cite{guo2015semantically}}. \textbf{KG2E} \cite{he2015learning} is a probabilistic embedding method for modeling the uncertainty in knowledge base. Notably, translation-based models demonstrate the state-of-the-art performance.

\subsection{Other Methods}
The \textbf{Unstructured Model (UM)} \cite{bordes2012joint} is a simplified version of TransE, which ignores relation information and the score function is reduced to $f_r(h,t) = ||\mathbf{h - t}||_2^2$. The \textbf{Structured Embedding (SE)} model \cite{bordes2011learning} transforms the entity space with the head-specific and tail-specific matrices and the score function is defined as $f_r(h,t)=||\mathbf{M_{h,r}h - M_{t,r}t}||$. The \textbf{Semantic Matching Energy (SME)} model \cite{bordes2012joint} \cite{bordes2014semantic} can enhance SE by considering the correlations between entities and relations with different matrix operators, as follows:
\begin{eqnarray}
f_r(h,t) = (\mathbf{M_1h+M_2r+b_1})^\top(\mathbf{M_3t+M_4r+b_2}) \nonumber \\
f_r(h,t) = (\mathbf{M_1h \otimes M_2r+b_1})^\top(\mathbf{M_3t \otimes M_4r+b_2}) \nonumber
\end{eqnarray}
where $\mathbf{M_1, M_2, M_3}$ and $\mathbf{M_4}$ are weight matrices, $\otimes$ is the Hadamard product, $\mathbf{b_1}$ and $\mathbf{b_2}$ are bias vectors. The \textbf{Single Layer Model (SLM)} applies neural network to knowledge graph embedding and the score function is defined as $f_r(h,t)=\mathbf{u_r^\top}g(\mathbf{M_{r,1}h + M_{r,2}t}) $ where
$\mathbf{M_{r,1}, M_{r,2}}$ are relation-specific weight matrices. The \textbf{Latent Factor Model (LFM)}  \cite{jenatton2012latent}, \cite{sutskever2009modelling} makes use of the second-order correlations between entities by a quadratic form and the score function is as $f_r(h,t) = \mathbf{h^\top W_rt}$. The \textbf{Neural Tensor Network (NTN)} model \cite{socher2013reasoning} defines a very expressive score function to combine the SLM and LFM: $f_r(h,t) =\mathbf{u_r^\top}g(\mathbf{h^\top W_{\cdot \cdot r}t + M_{r,1}h + M_{r,2}t + b_r}) $, where $\mathbf{u_r}$ is a relation-specific linear layer, $g(\cdot)$ is the $tanh$ function, $\mathbf{W} \in \mathbb{R}^{d \times d \times k}$ is a 3-way tensor. Besides, \textbf{RESCAL} is a collective matrix factorization model which is also a common method in knowledge graph embedding \cite{nickel2011three}, \cite{nickel2012factorizing}.

\section{Methods}
In this section, we introduce the novel manifold-based principle and then we analyze these methods from the algebraic and geometric perspectives.

\subsection{ManifoldE : a Manifold-Based Embedding Model}
Instead of adopting the translation-based principle $\mathbf{h+r=t}$, we apply the manifold-based principle $\mathbf{\mathcal{M}(h, r, t)} = D_r^2$ for a specific triple $(h,r,t)$. When a head entity and a relation are given, the tail entities lay in a high-dimensional manifold. Intuitively, our score function is designed by measuring the distance of the triple away from the manifold:
\begin{eqnarray}
f_r(h,t) = ||\mathcal{M}(h, r, t) - D_r^2 ||^2
\end{eqnarray}
where 
%$\mathbf{h,r,t}$ are respectively the embeddings of %the fact.
%\textit{h, r, and t}.
$D_r$ is a relation-specific manifold parameter. $\mathcal{M} : \mathbb{E} \times \mathbb{L} \times \mathbb{E} \xrightarrow{} \mathbb{R}$ is the manifold function, where $\mathbb{E}$, $\mathbb{L}$ are the entity set and relation set and $\mathbb{R}$ is the real number field.

\textbf{Sphere.} Sphere is a very typical manifold. In this setting, all the tail (or head) entities for a specific fact such as $\mathrm{(h,r,*)}$ are supposed to lay in a high-dimensional sphere where $\mathbf{h+r}$ is the center and $D_r$ is the radius, formally stated as below:
$$\mathcal{M}(h,r,t) = ||\mathbf{h+r-t}||_2^2$$

Obviously, this is a straight-forward extension of translation-based models in which  $D_r$ is zero. From the geometric perspective, the manifold collapses into a point when applying the translation-based principle.

Reproducing Kernel Hilbert Space (RKHS) usually provides a more expressive approach to represent the manifolds, which motivates us to apply the manifold-based principle with kernels.  To this point, kernels are involved to lay the sphere in a Hilbert space (an implicit high-dimensional space) as below:
\begin{eqnarray}
\mathcal{M}(h,r,t) & = & ||\varphi(h) + \varphi(r) - \varphi(t)||^2 \nonumber \\
& =  &\mathbf{K}(h, h) + \mathbf{K}(t, t) + \mathbf{K}(r, r) \\
& & - 2 \mathbf{K}(h, t) - 2 \mathbf{K}(r, t) + 2 \mathbf{K}(r, h) \nonumber
\end{eqnarray}
where $\varphi$ is the mapping from the original space to the Hilbert space, and $\mathbf{K}$ is the induced kernel by $\varphi$. Commonly, $\mathbf{K}$ could be Linear kernel ($\mathbf{K}(a,b) =\mathbf{a^\top b}$), Gaussian kernel ($\mathbf{K}(a,b) = e^{-\frac{||\mathbf{a-b}||^2_2}{\sigma^2}}$), Polynomial Kernel ($\mathbf{K}(a,b)=(\mathbf{a^\top b} + d)^p$, and so on.
Obviously, if applying the linear kernel, the above function is reduced to the original sphere manifold.
\begin{table}
	\centering
	\caption{Statistics of datasets}
	\label{tab1}
	\renewcommand\arraystretch{1.2}
	\begin{tabular}{|m{0.15\linewidth}<{\centering}|m{0.14\linewidth}<{\centering}|m{0.14\linewidth}<{\centering}|m{0.14\linewidth}<{\centering}|m{0.14\linewidth}<{\centering}|}
		\hline \textbf{Data} & \textbf{WN18} & \textbf{FB15K} & \textbf{WN11} & \textbf{FB13} \\
		\hline
		\hline \#Rel & 18 & 1,345 & 11 & 13 \\
		\hline \#Ent & 40,943 & 14,951 & 38,696 & 75,043 \\
		\hline \#Train & 141,442 & 483,142 & 112,581 & 316,232\\
		\hline \#Valid & 5,000 & 50,000 & 2,609 & 5,908\\
		\hline \#Test & 5,000 & 59,071 & 10,544 & 23,733\\
		\hline
	\end{tabular}
\end{table}

\begin{table*}
	\centering
	\caption{Evaluation results on link prediction}
	\label{tab2}
	\renewcommand\arraystretch{1.05}
	\begin{tabular}{|*{10}{c|}}
		\hline \textbf{Datasets} & \multicolumn{4}{c|}{\textbf{WN18}}  &  \multicolumn{4}{c|}{\textbf{FB15K}}  \\
		\hline
		\hline
		\multirow{2}*{Metric} & \multicolumn{2}{c|}{HITS@10(\%)} & \multicolumn{1}{c|}{HITS@1(\%)} & Time(s)  & \multicolumn{2}{c|}{HITS@10(\%)} & \multicolumn{1}{c|}{HITS@1(\%)} & Time(s) \\
		\cline{2-9} & Raw & Filter & Filter & One Epos & Raw & Filter & Filter & One Epos\\
		\hline
		%Unstructured \cite{bordes2011learning} & 35.3 & 38.2 & - & - & 4.5 & 6.3 & - & -\\
		%RESCAL \cite{nickel2012factorizing} & - & 37.2 & 52.8 & - & 28.4 & 44.1 \\
		SE　\cite{bordes2011learning}& 68.5 & 80.5  & - & - & 28.8 & 39.8 &- & -\\
		%SME \cite{bordes2012joint} & 65.1 & 74.1 & - & - & 30.7 & 40.8 & -  & -\\
		%SME(bilinear) \cite{bordes2012joint} & - & 54.7 & 61.3 & - & 31.3 & 41.3  \\
		%LFM \cite{jenatton2012latent} & - & 71.4 & 81.6 & - & 26.0 & 33.1  \\
		TransE \cite{bordes2013translating}  & 75.4 & 89.2 & 29.5 & \textbf{0.4} & 34.9 & 47.1 & 24.4& \textbf{0.7} \\
		TransH \cite{wang2014knowledge} & 73.0 & 82.3 & 31.3 & 1.4 & 48.2 & 64.4 & 24.8 & 4.8 \\
		TransR \cite{lin2015learning} & 79.8 & 92.0 & 33.5 & 9.8 & 48.4 & 68.7 & 20.0 & 29.1 \\
		%PTransE \cite{lin2015modeling} & - & - & -  & - & 51.4 & 84.6 & 61.8 & 266.0 \\
		KG2E \cite{he2015learning} & 80.2 & 92.8 & 54.1 & 10.7 & 48.9 & 74.0 & 40.4 & 44.2 \\
		\hline \hline
		ManifoldE Sphere & 81.1 &\textbf{ 94.4} & 26.4  & \textbf{0.4}  & 52.0 & \textbf{79.5} & 50.2 & \textbf{0.7}  \\
		ManifoldE Hyperplane & \textbf{81.4} & 93.7 & \textbf{90.8} & 0.5 & \textbf{52.6} & 78.2 & \textbf{54.0} & 0.8 \\
		\hline
	\end{tabular}
\end{table*}

\begin{figure}
\centering
\includegraphics[width=1.0\linewidth]{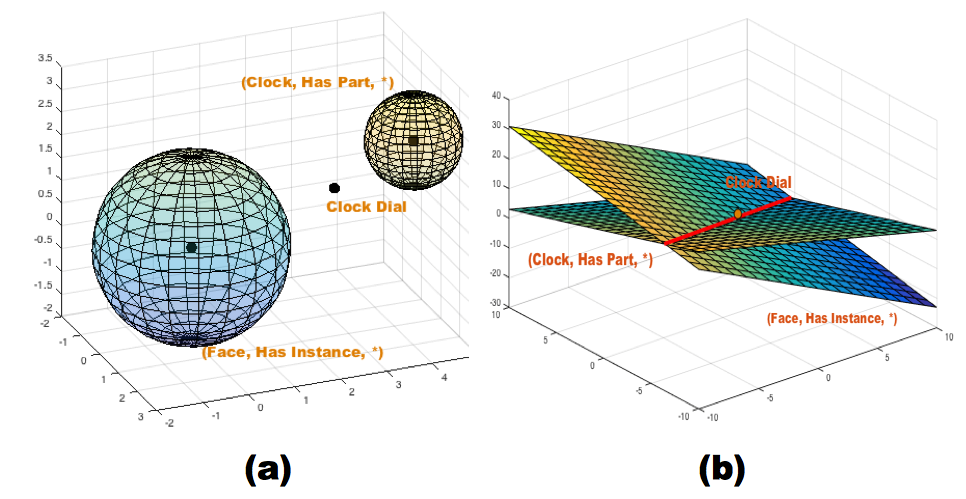}
\caption{Visualization of embedding for Manifold-based models. (a) corresponds to the Sphere setting where all the tail entities are supposed to lay in the sphere. As \textrm{Clock Dial} is matched by the two facts, it should lay in both spheres. (b) corresponds to the Hyperplane setting where \textrm{Clock Dial} should lay and does lay in both hyperplanes, making embedding more precise.}
\label{fig:fig_8}
\end{figure}

\textbf{Hyperplane.} As shown in Fig.\ref{fig:fig_8}, we could see that when two manifolds are not intersected, there may be a loss in embedding. Two spheres would intersect only under some strict conditions, while two hyperplanes would intersect if their normal vectors are not in parallel. Motivated by this fact, we apply a hyperplane to enhance our model as below: $$\mathcal{M}(h,r,t) = \mathbf{(h+r_{head})^\top(t+r_{tail})}$$ where $\mathbf{r_{head}}$ and $\mathbf{r_{tail}}$ are two specific relation embeddings. 
From the geometric perspective, given the head entity and the relation, the tail entities lay in the hyperplane whose direction is $\mathbf{h + r_{head}}$ and the bias corresponds to $D_r^2$. In practical cases, since the two vectors $\mathbf{e_1+r_{1,head}}$ and $\mathbf{e_2+r_{2,head}}$ are not likely to be parallel, there would be more chance to lead two intersected hyperplanes than two intersected spheres. Therefore, there would be more solutions provided by the intersection of hyperplanes.

Motivated by enlarging the number of precisely predicted tail entities for the same head and relation, we apply the absolute operators as $\mathcal{M}(h,r,t) = \mathbf{|h+r_{head}|^\top|t+r_{tail}|}$ where $|\mathbf{w}| \doteq (|w_1|, |w_2|, |w_3|, ... ,|w_n|)$. For an instance of one-dimensional case that $|h+r_{head}||t+r_{tail}|=D_r^2$, the absolute operator would double the solution number of $t$, meaning that two tail entities rather than one could be matched precisely to this head for this relation. For this reason, the absolute operator would promote the flexibility of embedding.

We also apply the kernel trick to the Hyperplane setting, as below: $$\mathcal{M}(h,r,t) = \mathbf{\mathbf{K}(h+r_{head}, t+r_{tail})}$$

\subsection{Algebraic Perspective}
The ill-posed equation system that posses more equations than free variables always leads to some undesired properties such as instability, which may be the reason why the translation-based principle performs not so well in precise link prediction. To alleviate this issue, manifold-based methods model embedding within a nearly well-posed algebraic framework, \textit{since our principle indicates only one equation for one fact triple}. Taking an example of sphere as $\sum_{i=1}^{d} (h_i+r_i-t_i)^2 = D_r^2$, we could conclude that if $d \ge \frac{\#Equation}{E+R} = \frac{T}{E+R}$, our embedding system would be more algebraically stable and this condition is easy to achieve by just enlarging the embedding dimension to a suitable degree. In theory, larger embedding dimension provides more solutions to embedding equations, which makes embedding more flexible. When the suitable condition is satisfied, the stable algebraic solution would lead the embedding to a fine characterization, therefore the precise link prediction would be promoted.

\subsection{Geometric Perspective}
The translation-based principle allocates just one position for a golden triple. We extend one point to a whole manifold such as a high dimensional sphere. For instance, all tail entities for a 1-N relation could lay on a sphere, which applies $\mathbf{h+r}$ as the center and $D_r$ as the radius. Obversely, it would be more suitable in a manifold setting than in a point setting. %With more precise fitting for the knowledge graph in manifold-based principle, the precise link prediction would be improved satisfactorily.

\subsection{Training}
We train our model with the rank-based hinge loss, which means to maximize the discriminative margin between the golden triples and the false ones.
\begin{eqnarray}
\mathcal{L}	 = \sum_{(h,r,t) \in \Delta} \sum_{(h',r',t') \in \Delta'} [f_{r'}(h',t') - f_r(h,t) + \gamma]_{+}
%& & + \lambda \left( \sum_{e \in E} ||\mathbf{e}||_2^2 + \sum_{r \in R} ||\mathbf{r}||^2_2 + \sum_{r \in R} ||D_r||^2 %\right)
\end{eqnarray}
where $\mathcal{L}$ is the loss function which should be minimized, $\gamma$ is the margin, and $[\cdot]_{+} \doteq \max(0, \cdot)$ is the hinge loss. The false triples are sampled with the Bernoulli Sampling Method as introduced in \cite{wang2014knowledge}. We initialize the embedding vectors by the similar methods used in deep neural network \cite{glorot2010understanding}. Stochastic gradient descent (SGD) is applied to solve this problem.

In theory, our computation complexity relative to TransE is bounded by a very small constant, as $O(\lambda \times O(\mathrm{TransE}))$ where $\lambda \ge 1$. This small constant $\lambda$ is caused by manifold-based operations and kernelization. Commonly, TransE is the most efficient among all the translation-based methods, while ManifoldE could be comparable to TransE in efficiency, hence faster than other translation-based methods.

\begin{table*}
	\centering
	\caption{Evaluation results on FB15K by mapping properties of relations(\%)}
	\label{tab3}
	\renewcommand\arraystretch{1.05}
	\begin{tabular}{|*{9}{c|}}
		\hline \textbf{Tasks} & \multicolumn{4}{c|}{\textbf{Predicting Head(HITS@10)}}  &  \multicolumn{4}{c|}{\textbf{Predicting Tail(HITS@10)}}  \\
		\hline
		\hline
		Relation Category & 1-1 & 1-N & N-1 & N-N & 1-1 & 1-N & N-1 & N-N \\
		\hline
		%Unstructured \cite{bordes2011learning} & 34.5 & 2.5 & 6.1 & 6.6 & 34.3 & 4.2 & 1.9 & 6.6 \\
		%SE　\cite{bordes2011learning} & 35.6 & 62.6 &17.2 & 37.5& 34.9 & 14.6 & 68.3 & 41.3  \\
		%SME \cite{bordes2012joint} & 35.1 & 53.7& 19.0 & 40.3 & 32.7& 14.9 & 61.6 & 43.3 \\
		%SME \cite{bordes2012joint} & 30.9 & 69.6 & 19.9 & 38.6 & 28.2 & 13.1 & 76.0 & 41.8 \\
		TransE \cite{bordes2013translating} & 43.7 &65.7 & 18.2 & 47.2 & 43.7 & 19.7 & 66.7 & 50.0 \\
		TransH \cite{wang2014knowledge} & 66.8 & 87.6 & 28.7 & 64.5 & 65.5 & 39.8 & 83.3 & 67.2 \\
		TransR \cite{lin2015learning} & 78.8 & 89.2 & 34.1 & 69.2 & 79.2 & 37.4 & 90.4 & 72.1 \\
		%PTransE \cite{lin2015modeling} & 90.1 & 92.0 & 58.7 & 86.1 & 90.1 & 70.7 & 87.5 & 88.7 \\
		%KG2E \cite{he2015learning} & 92.3 & 93.7 & \textbf{66.0} & 69.6 & 92.6 & 67.9 & 94.4 & 73.4 \\
		% TransA & 79.9 & \textbf{93.9} & 33.1 & \textbf{74.3} & 79.8 & \textbf{40.1} & 93.2 & \textbf{77.1} \\
		\hline \hline
		ManifoldE Sphere & \textbf{80.9} & 94.5 & 36.7 & \textbf{82.1} & \textbf{80.0} & \textbf{50.7} & 93.6 & 82.5 \\
		ManifoldE Hyperplane & 79.0 & \textbf{94.9} & \textbf{38.4} & 80.2 & 76.7 & 49.5 & \textbf{94.2} & \textbf{83.0} \\
		\hline
	\end{tabular}
\end{table*}

\begin{table*}
	\centering
	\caption{Evaluation results for precise prediction on FB15K by mapping properties of relations(\%)}
	\label{tab5}
	\renewcommand\arraystretch{1.05}
	\begin{tabular}{|*{9}{c|}}
		\hline \textbf{Tasks} & \multicolumn{4}{c|}{\textbf{Predicting Head(HITS@1)}}  &  \multicolumn{4}{c|}{\textbf{Predicting Tail(HITS@1)}}  \\
		\hline
		\hline
		Relation Category & 1-1 & 1-N & N-1 & N-N & 1-1 & 1-N & N-1 & N-N \\
		\hline
		TransE \cite{bordes2013translating} & 35.4 & 50.7 & 8.6 & 18.1 & 34.5 & 10.6 & 56.1 & 20.3 \\
		TransH \cite{wang2014knowledge} & 35.3 & 48.7 & 8.4 & 16.9 & 35.5 & 10.4 & 57.5 & 19.3 \\
		TransR \cite{lin2015learning} & 29.5 & 42.8 & 6.1 & 14.5 & 28.0 & 7.7 & 44.1 & 16.2  \\
		%PTransE \cite{lin2015modeling} & 57.5 & 83.0 & 46.2 & 60.3 & 58.1 & \textbf{58.4} & 73.9 & 61.7 \\
		%KG2E \cite{he2015learning} & 62.3 & 73.9 & 39.4 & 30.4 & 62.3 & 33.9 & 76.2 & 33.8 \\
		\hline \hline
		ManifoldE Sphere & 33.3 & 60.0 & 13.3 & \textbf{53.8} & 34.5 & 19.5 & 58.5 & \textbf{56.2} \\
		ManifoldE Hyperplane &\textbf{ 62.6} & \textbf{78.6} & \textbf{21.5} & 53.0 & \textbf{58.2} & \textbf{32.1} & \textbf{77.8} & 55.9 \\
		\hline
	\end{tabular}
\end{table*}

\section{Experiments}
Our experiments are conducted on four public benchmark datasets that are the subsets of Wordnet \cite{miller1995wordnet} and Freebase \cite{bollacker2008freebase}. The statistics of these datasets are listed in Tab.\ref{tab1}. Experiments are conducted on two tasks : Link Prediction and Triple Classification. To further demonstrate how the proposed model performs the manifold-based principle, we present the visualization comparison between translation-based and manifold-based models in the section 4.3. Finally, we conduct error analysis to further understand the benefit and limits of our models.

\subsection{Link Prediction}
Reasoning is the focus of knowledge computation. To verify the reasoning performance of embedding, link prediction task is conducted. This task aims to predict the missing entities. An alternative of the entities and the relation are given when the embedding methods infer the other missing entity. More specifically, in this task, we predict $t$ given $(h, r, *)$, or predict $h$ given $(*,  r, t)$ . The WN18 and FB15K are two benchmark datasets for this task. Notably, many AI tasks could be enhanced by ``Link Prediction'', such as relation extraction \cite{hoffmann2011knowledge}.

\textbf{Evaluation Protocol.} We adopt the same protocol used in previous studies. Firstly, for each testing triple $(h,r,t)$, we corrupt it by replacing the tail $t$ (or the head $h$) with every entity $e$ in the knowledge graph. Secondly, we calculate a probabilistic score of this corrupted triple with the score function $f_r(h,t)$. By ranking these scores in descending order, we then obtain the rank of the original triple. The evaluation metric is the proportion of testing triple whose rank is not larger than N (HITS@N). HITS@10 is applied for common reasoning ability and HITS@1 concerns the precise embedding performance. This is called ``Raw'' setting. When we filter out the corrupted triples that exist in the training, validation, or test datasets, this is the``Filter'' setting. If a corrupted triple exists in the knowledge graph, ranking it ahead the original triple is also acceptable. To eliminate this case, the ``Filter'' setting is more preferred. In both settings, a higher HITS@N means better performance. Note that we do not report the results of ``raw'' setting for HITS@1, because they are too small to make a sense. Notably, we actually run each baseline in the same setting for five times, and average the running time as the results. 

\textbf{Implementation.} As the datasets are the same, we directly reproduce the experimental results of several baselines from the literature for HITS@10. As to HITS@1, we request the results from the authors of PTransE and KG2E. We acknowledge these authors \textit{Yankai Lin} and \textit{Shizhu} He. We have attempted several settings on the validation dataset to get the best configuration. Under the ``bern.'' sampling strategy, the optimal configurations of ManifoldE are as follows. For sphere, $\alpha = 0.001$, $k=100$, $\gamma = 3.0$, Linear kernel, on WN18; $\alpha = 0.0005$, $k=800$, $\gamma = 1.0$, Polynomial kernel($p=2, d=2$) on FB15K. For hyperplane, learning rate $\alpha=0.01$, embedding dimension $k=800$, margin $\gamma=0.2$, Linear kernel, on WN18; $\alpha=0.01$, $k=1000$, $\gamma=0.2$, Linear kernel, on FB15K. The experimental environment is a common PC with i7-4790 CPU, 16G Memory and Windows 10. Note that all the symbols are introduced in ``Methods''. Notably, we train the model until convergence about 10,000 rounds in previous version. But in this version, we adopt no trick and train the model until 2,000 rounds.

\textbf{Results.} Evaluation results on WN18 and FB15K are reported in Tab.\ref{tab2} and Tab.\ref{tab3}. We observe that:
\begin{enumerate}
	\item ManifoldE beats all the baselines in all the sub-tasks, yielding the effectiveness and efficiency of the manifold-based principle.
	\item From the algebraic perspective, it's reasonable to measure the algebraic ill-posed degree with the radio $T/(E+R)$ because with the translation-based principle, $Td$ is the number of equations and $(E+R)d$ is the number of free variables, a larger radio means more ill-posed. Since the manifold-based principle alleviates this issue, ManifoldE(Sphere) would make a more promotion relatively to the comparable baselines (TransE) under a larger radio. As to the metric HITS@1, on WN18, the radio is 3.5 while TransE achieves 29.5\% and ManifoldE(Sphere) achieves 55\%, leading to a relative improvement of 85.1\%. On FB15K the radio is 30.2 while TransE achieves 24.4\% and ManifoldE(Sphere) achieves 64.1\%, leading to a relative improvement of 162.7\%. This comparison illustrates manifold-based methods could stabilize the algebraic property of embedding system, by which means, the precise embedding could be approached much better.
	\item From the geometric perspective, traditional models attempt to express all the matched entities into one position, which leads to unsatisfactory performance on complex relations. Meanwhile, manifold-based model could perform much better for these complex relations as we discussed. As to the metric HITS@1, the simple relation 1-1 improves relatively by 87.8\% by ManifoldE(Sphere) than TransE while the complex relations such as 1-N, N-1, N-N improve relatively by 266.5\%, 36.2\% and 215.7\% respectively. This comparison demonstrates manifold-based method that extends the golden position from one point to a manifold could better characterize the true facts, especially for complex relations.
%	\item ManifoldE also achieves consistent and significant improvements on the Relation Link Prediction as expected. These results also prove the effectiveness of manifold-based principle that alleviates the unsatisfactory precise prediction issue to a big extent.
\end{enumerate}

\subsection{Triple Classification}
In order to present the discriminative capability of our method between true and false facts, triple classification is conducted. This is a classical task in knowledge base embedding, which aims at predicting whether a given triple $(h,r, t)$ is correct or not. WN11 and FB13 are the benchmark datasets for this task. Note that evaluation of classification needs negative samples, and the datasets have already been built with negative triples.

\begin{table}
	\centering
	\caption{Triple classification: accuracy(\%) for different embedding methods.}
	\label{tab4}
	\renewcommand\arraystretch{1.1}
	\begin{tabular}{|c|c|c|c|}
		\hline \textbf{Methods} & \textbf{WN11} & \textbf{FB13} & \textbf{AVG.} \\
		\hline
		\hline
		SE & 53.0 & 75.2 & 64.1 \\
		%SME(bilinear) & 70.0 & 63.7 & 66.9\\
		%LFM  & 73.8 & 84.3 & 79.0\\
		NTN  & 70.4 & 87.1 & 78.8 \\
		TransE & 75.9 & 81.5 & 78.7 \\
		TransH & 78.8 & 83.3 & 81.1 \\
		TransR & 85.9 & 82.5 & 84.2\\
		KG2E & 85.4 & 85.3 & 85.4 \\
		\hline \hline
		ManifoldE Sphere & \textbf{87.5} & 87.2 & \textbf{87.4} \\
		ManifoldE Hyperplane & 86.9 & \textbf{87.3} & 87.1 \\	
		\hline
	\end{tabular}
\end{table}

\textbf{Evaluation Protocol.} The decision process is very simple as follows: for a triple $(h,r,t)$, if $f_r(h,t)$ is below a threshold $\sigma_r$, then positive; otherwise negative. The thresholds $\{\sigma_r\}$ are determined on the validation dataset. This task is somehow a triple binary classification.

\textbf{Implementation.} As all methods use the same datasets,  we directly re-use the results of different methods from the literature. We have attempted several settings on the validation dataset to find the best configuration. The optimal configurations of ManifoldE are as follows with ``bern'' sampling. For sphere, $\alpha = 0.001$, $k=100$, $\gamma = 10.0$, Linear kernel on WN18; $\alpha=0.00005$, $k=1000$, $\gamma=0.3$, Gaussian kernel ($\sigma=1.0$) on FB13. For hyperplane, learning rate $\alpha=0.01$, embedding dimension $k=500$, margin $\gamma=1.0$, Linear kernel, on WN18; $\alpha=0.001$, $k=1000$, $\gamma=3.0$, Polynomial kernel ($p=2, d=2$), on FB13.

\textbf{Results.} Accuracies are reported in Tab.\ref{tab4}. We observe that:
\begin{enumerate}
	\item Overall, ManifoldE yields the best performance. This illustrates our method could improve the embedding.
	\item More specifically, on WN11, the relation ``Type Of'' that is a complex one, improves from 71.4\% of TransE to 86.3\% of ManifoldE(Sphere) while on FB13, the relation ``Gender'' that is an extreme N-1 relation, improves from 95.1\% to 99.5\%. This comparison shows manifold-based methods could handle complex relations better.
\end{enumerate}

\subsection{Visualization Comparison between Translation-Based and Manifold-Based Principle}
As the Fig.\ref{fig:fig_1} shows, the translation-based principle involves too much noise near the center where is supposed to lay the true facts. We attribute such issues to the precise link prediction issue as introduced previously, However, manifold-based principle alleviates this issue to enhance precise knowledge embedding, which could be seen from the visualization results.

%\subsection{Case Study}
%To analyze the improvement of precise link prediction in a fine-grained level, we sample some triples which fail in TransR but are well dealt with in ManifoldE (Hyperplane). Then, we present two classical examples. First, for the testing triple $\mathrm{(East~Timor, Part~Of, Southest~Asia)}$ whose rank is 7, the top tails are $\mathrm{Europe}$, $\mathrm{Hawaiian}$, $\mathrm{Haiti}$, $\mathrm{Caribbean}$, $\mathrm{Greater~Antilles}$ and $\mathrm{Polynesia}$ which are all about confusing locations. Next, for the triple $\mathrm{(Competition, Has~Instance, Match)}$ whose rank is 4, the top tails are $\mathrm{Social~Event}$, $\mathrm{Final}$ and $\mathrm{Battle}$, which are all competition-related. In conclusion, ManifoldE could distinguish similar concepts to promote precise knowledge prediction.

\subsection{Error Analysis}
To analyze the errors in Link Prediction, we randomly sample 100 testing triples that could not rank at the top positions by ManifoldE (Hyperplane) and three categories of errors are summarized. Notably, we call the predicted top rank triple, which is not the golden one, as ``top rank triple''.
\begin{enumerate}
	\item \textbf{True Facts (29\%)}: The top rank triple is correct though it is not contained in the knowledge graph, thus ranking it before the golden one is acceptable. This category is caused by incompleteness of the dataset. For example, reflexive semantics as $\mathrm{(Hot, Similar~To, Hot)}$, general expression as $\mathrm{(Animal~Scientist, Type~Of, Individual)}$, professional knowledge as $\mathrm{(Crescentia, Type~Of, Plant)}$ and so on.
	\item \textbf{Related Concepts (63\%)}: The top rank triple is a related concept, but the corresponding fact is not exactly correct. This category is caused by the relatively simple manifolds applied by ManifoldE. For example, puzzled place membership as $\mathrm{(Afghanistan, Has~Part, Hyderabad)}$, similar mentions as $\mathrm{(New~Haven, Part~Of, One~Star~State)}$, similar concepts as $\mathrm{(Sea , Has~Instance, Aleutian~Islands)}$, possible knowledge as $\mathrm{(Accent, Type~Of, Individual)}$ and so on. We could further exploit complex manifolds to enhance the discriminative ability.
	\item \textbf{Others (8\%)}:
	There are always some top rank triples that are difficult to interpret. 
	%For example, $\mathrm{(Demonstration, Type~Of, Body)}$, $\mathrm{(Geographic~Area, Has~Instance, Argun)}$ and so on.
\end{enumerate}

\section{Conclusions}
In this paper, we study the precise link prediction problem and reveal two reasons to the problem: the ill-posed algebraic system and the over-restricted geometric form. To alleviate these issues, we propose a novel manifold-based principle and the corresponding ManifoldE models~(Sphere/Hyperplane) inspired by the principle. From the algebraic perspective, ManifoldE is a nearly well-posed equation system and from the geometric perspective, it expands point-wise modeling in the translation-based principle to manifold-wise modeling. Extensive experiments show our method achieves substantial improvements against the state-of-the-art baselines.

This work was partly supported by the National Basic Research Program (973 Program) under grant No.2012CB316301 / 2013CB329403, and the National Science Foundation of China under grant No.61272227 / 61332007.

%To reproduce our result, experimental codes and data will be published in github\footnote{http://www.github.com}.

%We list some future work:
%\begin{itemize}
%	\item We would explore different manifolds rather than high-dimensional spheres to enhance the embedding. As the experiments argued, this work would make much sense.
%	\item We would involve the entities vector transformation, which are applied for translation-based principle such as TransR and TransH, to manifold-based principle.
%\end{itemize}

\bibliography{ijcai}

\begin{thebibliography}{}

\bibitem[\protect\citeauthoryear{Bollacker \bgroup \em et al.\egroup
  }{2008}]{bollacker2008freebase}
Kurt Bollacker, Colin Evans, Praveen Paritosh, Tim Sturge, and Jamie Taylor.
\newblock Freebase: a collaboratively created graph database for structuring
  human knowledge.
\newblock In {\em Proceedings of the 2008 ACM SIGMOD international conference
  on Management of data}, pages 1247--1250. ACM, 2008.

\bibitem[\protect\citeauthoryear{Bordes \bgroup \em et al.\egroup
  }{2011}]{bordes2011learning}
Antoine Bordes, Jason Weston, Ronan Collobert, Yoshua Bengio, et~al.
\newblock Learning structured embeddings of knowledge bases.
\newblock In {\em Proceedings of the Twenty-fifth AAAI Conference on Artificial
  Intelligence}, 2011.

\bibitem[\protect\citeauthoryear{Bordes \bgroup \em et al.\egroup
  }{2012}]{bordes2012joint}
Antoine Bordes, Xavier Glorot, Jason Weston, and Yoshua Bengio.
\newblock Joint learning of words and meaning representations for open-text
  semantic parsing.
\newblock In {\em International Conference on Artificial Intelligence and
  Statistics}, pages 127--135, 2012.

\bibitem[\protect\citeauthoryear{Bordes \bgroup \em et al.\egroup
  }{2013}]{bordes2013translating}
Antoine Bordes, Nicolas Usunier, Alberto Garcia-Duran, Jason Weston, and Oksana
  Yakhnenko.
\newblock Translating embeddings for modeling multi-relational data.
\newblock In {\em Advances in Neural Information Processing Systems}, pages
  2787--2795, 2013.

\bibitem[\protect\citeauthoryear{Bordes \bgroup \em et al.\egroup
  }{2014}]{bordes2014semantic}
Antoine Bordes, Xavier Glorot, Jason Weston, and Yoshua Bengio.
\newblock A semantic matching energy function for learning with
  multi-relational data.
\newblock {\em Machine Learning}, 94(2):233--259, 2014.

\bibitem[\protect\citeauthoryear{Fan \bgroup \em et al.\egroup
  }{2014}]{fan2014transition}
Miao Fan, Qiang Zhou, Emily Chang, and Thomas~Fang Zheng.
\newblock Transition-based knowledge graph embedding with relational mapping
  properties.
\newblock In {\em Proceedings of the 28th Pacific Asia Conference on Language,
  Information, and Computation}, pages 328--337, 2014.

\bibitem[\protect\citeauthoryear{Glorot and
  Bengio}{2010}]{glorot2010understanding}
Xavier Glorot and Yoshua Bengio.
\newblock Understanding the difficulty of training deep feedforward neural
  networks.
\newblock In {\em International conference on artificial intelligence and
  statistics}, pages 249--256, 2010.

\bibitem[\protect\citeauthoryear{Guo \bgroup \em et al.\egroup
  }{2015}]{guo2015semantically}
Shu Guo, Quan Wang, Bin Wang, Lihong Wang, and Li~Guo.
\newblock Semantically smooth knowledge graph embedding.
\newblock In {\em Proceedings of ACL}, 2015.

\bibitem[\protect\citeauthoryear{He \bgroup \em et al.\egroup
  }{2015}]{he2015learning}
Shizhu He, Kang Liu, Guoliang Ji, and Jun Zhao.
\newblock Learning to represent knowledge graphs with gaussian embedding.
\newblock In {\em Proceedings of the 24th ACM International on Conference on
  Information and Knowledge Management}, pages 623--632. ACM, 2015.

\bibitem[\protect\citeauthoryear{Hoffmann \bgroup \em et al.\egroup
  }{2011}]{hoffmann2011knowledge}
Raphael Hoffmann, Congle Zhang, Xiao Ling, Luke Zettlemoyer, and Daniel~S Weld.
\newblock Knowledge-based weak supervision for information extraction of
  overlapping relations.
\newblock In {\em Proceedings of the 49th Annual Meeting of the Association for
  Computational Linguistics: Human Language Technologies-Volume 1}, pages
  541--550. Association for Computational Linguistics, 2011.

\bibitem[\protect\citeauthoryear{Jenatton \bgroup \em et al.\egroup
  }{2012}]{jenatton2012latent}
Rodolphe Jenatton, Nicolas~L Roux, Antoine Bordes, and Guillaume~R Obozinski.
\newblock A latent factor model for highly multi-relational data.
\newblock In {\em Advances in Neural Information Processing Systems}, pages
  3167--3175, 2012.

\bibitem[\protect\citeauthoryear{Ji \bgroup \em et al.\egroup }{}]{JiKnowledge}
Guoliang Ji, Shizhu He, Liheng Xu, Kang Liu, and Jun Zhao.
\newblock Knowledge graph embedding via dynamic mapping matrix.

\bibitem[\protect\citeauthoryear{Lin \bgroup \em et al.\egroup
  }{2015a}]{lin2015modeling}
Yankai Lin, Zhiyuan Liu, and Maosong Sun.
\newblock Modeling relation paths for representation learning of knowledge
  bases.
\newblock {\em Proceedings of the 2015 Conference on Empirical Methods in
  Natural Language Processing (EMNLP). Association for Computational
  Linguistics}, 2015.

\bibitem[\protect\citeauthoryear{Lin \bgroup \em et al.\egroup
  }{2015b}]{lin2015learning}
Yankai Lin, Zhiyuan Liu, Maosong Sun, Yang Liu, and Xuan Zhu.
\newblock Learning entity and relation embeddings for knowledge graph
  completion.
\newblock In {\em Proceedings of the Twenty-Ninth AAAI Conference on Artificial
  Intelligence}, 2015.

\bibitem[\protect\citeauthoryear{Miller}{1995}]{miller1995wordnet}
George~A Miller.
\newblock Wordnet: a lexical database for english.
\newblock {\em Communications of the ACM}, 38(11):39--41, 1995.

\bibitem[\protect\citeauthoryear{Nickel \bgroup \em et al.\egroup
  }{2011}]{nickel2011three}
Maximilian Nickel, Volker Tresp, and Hans-Peter Kriegel.
\newblock A three-way model for collective learning on multi-relational data.
\newblock In {\em Proceedings of the 28th international conference on machine
  learning (ICML-11)}, pages 809--816, 2011.

\bibitem[\protect\citeauthoryear{Nickel \bgroup \em et al.\egroup
  }{2012}]{nickel2012factorizing}
Maximilian Nickel, Volker Tresp, and Hans-Peter Kriegel.
\newblock Factorizing yago: scalable machine learning for linked data.
\newblock In {\em Proceedings of the 21st international conference on World
  Wide Web}, pages 271--280. ACM, 2012.

\bibitem[\protect\citeauthoryear{Socher \bgroup \em et al.\egroup
  }{2013}]{socher2013reasoning}
Richard Socher, Danqi Chen, Christopher~D Manning, and Andrew Ng.
\newblock Reasoning with neural tensor networks for knowledge base completion.
\newblock In {\em Advances in Neural Information Processing Systems}, pages
  926--934, 2013.

\bibitem[\protect\citeauthoryear{Sutskever \bgroup \em et al.\egroup
  }{2009}]{sutskever2009modelling}
Ilya Sutskever, Joshua~B Tenenbaum, and Ruslan Salakhutdinov.
\newblock Modelling relational data using bayesian clustered tensor
  factorization.
\newblock In {\em Advances in neural information processing systems}, pages
  1821--1828, 2009.

\bibitem[\protect\citeauthoryear{Tikhonov and
  Arsenin}{1978}]{Tikhonov1978Solutions}
A.~N. Tikhonov and V.~Y. Arsenin.
\newblock Solutions of ill-posed problems.
\newblock {\em Mathematics of Computation}, 32(144):491--491, 1978.

\bibitem[\protect\citeauthoryear{Wang \bgroup \em et al.\egroup
  }{2014}]{wang2014knowledge}
Zhen Wang, Jianwen Zhang, Jianlin Feng, and Zheng Chen.
\newblock Knowledge graph embedding by translating on hyperplanes.
\newblock In {\em Proceedings of the Twenty-Eighth AAAI Conference on
  Artificial Intelligence}, pages 1112--1119, 2014.

\bibitem[\protect\citeauthoryear{Wang \bgroup \em et al.\egroup
  }{2015}]{wang2015knowledge}
Quan Wang, Bin Wang, and Li~Guo.
\newblock Knowledge base completion using embeddings and rules.
\newblock In {\em Proceedings of the 24th International Joint Conference on
  Artificial Intelligence}, 2015.

\end{thebibliography}
\bibliographystyle{ijcai}

\end{document}